\documentclass[runningheads]{llncs}

\usepackage{graphicx}
\usepackage{grffile}

\begin{document}

\title{Writer Independent Offline Signature Recognition Using Ensemble Learning}

\author{Sourya Dipta Das\inst{1} \and
Himanshu Ladia\inst{2} \and
Vaibhav Kumar\inst{2} \and
Shivansh Mishra\inst{2}}

\authorrunning{Sourya, Himanshu et al.}

\institute{Jadavpur University, Kolkata, India \and
Delhi Technological University, Delhi, India}

\maketitle              % typeset the header of the contribution

\begin{abstract}
The area of Handwritten Signature Verification has been broadly researched in the last decades, but remains an open research problem. In offline (static) signature verification, the dynamic information of the signature writing process is lost, and it is difficult to design good feature extractors that can distinguish genuine signatures and skilled forgeries. This verification task is even harder in writer independent scenarios which is undeniably fiscal for realistic cases. In this paper, we have proposed an Ensemble model for offline writer, independent signature verification task with Deep learning. We have used two CNNs for feature extraction, after that RGBT for classification \& Stacking to generate final prediction vector. We have done extensive experiments on various datasets from various sources to maintain a variance in the dataset. We have achieved the state of the art performance on various datasets.

\end{abstract}
\section{Introduction}
Signature verification is a critical task and many efforts have been made to remove the uncertainty involved in the manual authentication procedure, which makes it an important research line in the field of machine learning and pattern recognition. In offline signature verification, the signature is acquired after the writing process is completed, by scanning a document containing the signature. Defining discriminative feature extractors for offline signatures is a difficult task due to different types of signatures from different origin based on language, country or region . Offline signature verification can be addressed with writer dependent and writer independent approaches. Writer independent scenario is preferable over writer dependent approaches, as for a functioning system, a writer dependent system needs to be retrained with every new writer or signer. For a consumer based system, such as bank, where every day new consumers can open their account this incurs huge cost. Whereas, in writer independent case, a generic system is built to model the discrepancy among the genuine and forged signatures. One of the main challenges for the writer independent offline signature verification task is having a high intra-class variability compared to physical biometric traits, such as fingerprint or iris. This issue is aggravated with the presence of low inter-class variability when we consider skilled forgeries. In a realistic scenario, during training we only have access to genuine signatures for the users enrolled to the system. During operations, however, we want the system not only to be able to accept genuine signatures, but also to reject forgeries. Lastly, the amount of data available for each user is often very limited in real applications. During the enrollment phase, users are often required to supply only a few samples of their signatures. In other words, even if there is a large number of users enrolled to the system, a classifier needs to perform well for a new user, for whom only a small set of samples are available. In this scenario, our model is based on ensemble model of two deep nets which is trained on two different loss function to learn discriminative features for detection of both person \& forgery. With our model, we have achieved high accuracy on large scale cross domain datasets.

\section{Previous Work}
Different hand crafted features have been proposed for offline signature verification tasks. Examples include block codes, wavelet and Fourier series etc \cite{2}. Some other methods consider the geometrical and topological characteristics of local attributes, such as position, tangent direction, blob structure, connected component and curvature \cite{1}, grid based methods  \cite{7}, methods based on geometrical moments \cite{6}, and texture based features \cite{8}, Projection and contour based methods \cite{3} are also quite used for offline signature verification as a unsupervised method. Apart from that, some structural methods like graph matching \cite{4} and compact correlated features \cite{5} are also become popular in signature recognition. Convolutional Neural Network \cite{10} have been used to transfer learning of macro \& micro features from training on a large dataset to using this model as a feature extractor in a another task by using same network. From the perspective of few shot learning, Siamese networks are popular \& effective option for offline signature verification. In a recent work, Convolutional siamese network \cite{9} had been used for offline signature recognition.

\section{Our Model Description}
We have proposed a ensemble model consists of two deepnet model and we have used Regularized Gradient Boosting Tree (RGBT) as a classifier and finally stacking as a ensemble algorithm to get final prediction vector.

\subsection{Preprocessing \& Feature Extraction}
We have used preprocessing techniques from the paper \cite{10} \& then we have used two deep net models to extract the feature from each signature. One deepnet (which is called here as "Signet")  model is trained on genuine signatures to learn common descriptive features among the writers and another deepnet model (which is called here as "Signet-f") is trained on genuine \& forged signatures both to learn discriminative features between original and fake one. We pass the signature images through both the models and capture the output of the last layer of the convolutional neural network. This gives us features extracted from signet, and signetf models.

\subsection{Model Architecture}
We have used same Convolutional Neural Network Architecture mentioned in the paper \cite{10} for both CNNs and used their training procedure \& loss functions for respective CNNs on the same dataset to get trained CNNs for feature extraction. Description of the CNN is described in [Fig. 1].
%%\vspace{-0.7cm}

\begin{figure}
\centering
\includegraphics[scale = 0.3]{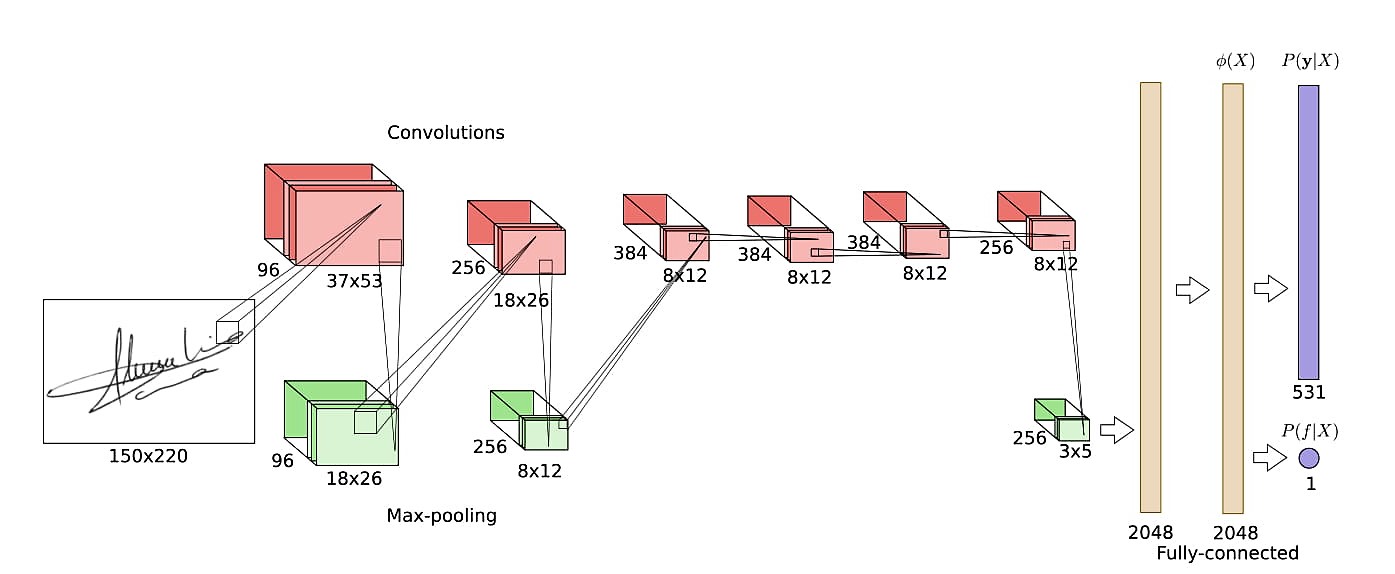}
\caption{Description of the CNN architecture used for feature extraction} \label{fig1}
\end{figure}

%%\vspace{-0.6cm}
\subsection{Writer Independent Classifier}
We have used Regularized Gradient Boosting Tree as our writer independent classifier. It is a sequential technique which works on the principle of an ensemble. It combines a set of weak learners and delivers improved prediction accuracy. At any instant t, the model outcomes are weighed based on the outcomes of previous instant t-1. The outcomes predicted correctly are given a lower weight and the ones miss-classified are weighted higher. Though fitting the training set too closely can lead to degradation of the model's generalization ability. So, to neutralize this effect, we have used  regularization methods to reduce this overfitting effect by constraining the fitting procedure. During training, we have set maximum tree depth for base learners as 3, boosting learning rate as 0.1 and number of boosted trees to fit as 100. After getting prediction vector from two classifier, they are fed to stacking module to form a ensemble.

\subsection{Stacking as Ensemble Method}
Stacking is one of the ensemble model, where a new model is trained from the combined predictions of two or more model. The predictions from the models are used as inputs for each sequential layer, and combined to form a final set of predictions. In the implementation, a logistic regression model is used as the combiner. 
%%\vspace{-0.3em}

\begin{figure}
\centering
\includegraphics[scale = 0.3]{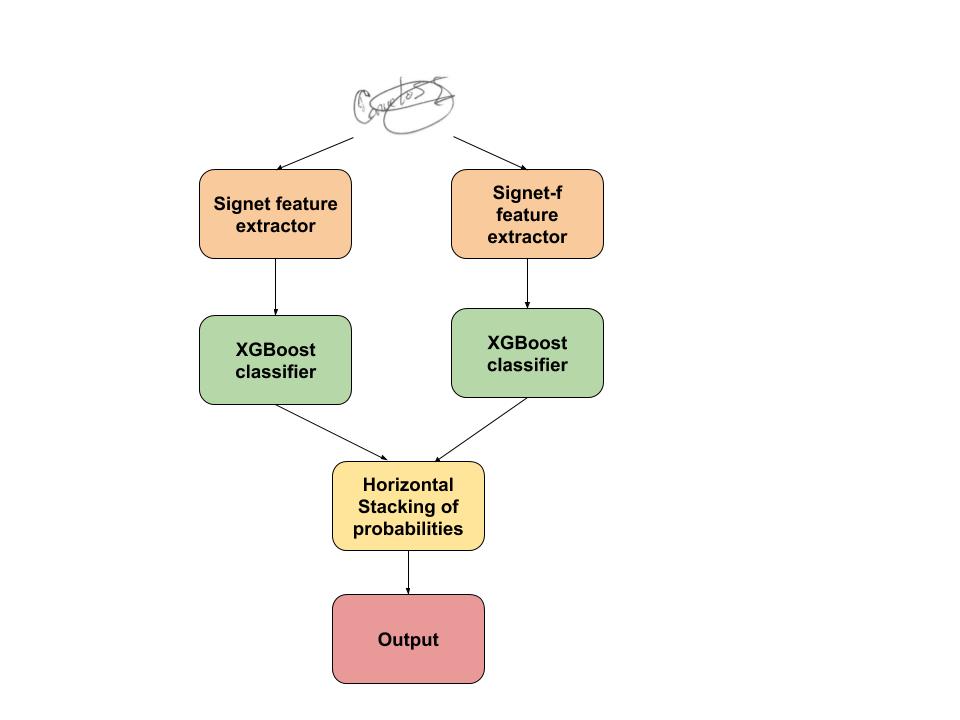}
\caption{Signature Verification Model Block Diagram} \label{fig1}
\end{figure}

%%\vspace{-0.4em}
\section{Experimentation}
The dataset combining of genuine and forged images is run through the CNN for extracting features. Then, a 33\% random test split is used. The 67\% data is used for training the RGBT model. Then, from the trained model, the probabilities is calculate for both, the 67\% training data and the 33\% testing data. The results are evaluated for both, signet and signet-f feature extractors. The corresponding probabilities are also calculated from the features extracted from a signet-f model in the similar fashion. Then, these probabilities are horizontally stacked to create a (n,2) shape feature matrix. This process is done for both, the 67\% training data, and 33\% test data. A logistic unit is trained on this feature matrix using the 67\% training data. And it is tested on the feature matrix of 33\% test data. For our implementation \& training , we had used a laptop with CPU - Intel(R) Core(TM) i5-6200U CPU @ 2.30GHz and RAM of 8 GB. During the experiments on various datsets, we observe that the signet model is better at classifying genuine images and the signet-f model is better at classifying forged images. So, it is logical to stack the two classifiers and improve the overall accuracy by a handsome margin.

\iffalse
\begin{table}[]
\centering
\begin{tabular}{|c|l|l|l|l|}
\hline
Dataset Name & \multicolumn{1}{c|}{\begin{tabular}[c]{@{}c@{}}Genuine\\ signatures\end{tabular}} & \multicolumn{1}{c|}{\begin{tabular}[c]{@{}c@{}}Fake\\ signatures\end{tabular}} & \multicolumn{1}{c|}{\begin{tabular}[c]{@{}c@{}}Samples in \\ Training Set\end{tabular}} & \multicolumn{1}{c|}{\begin{tabular}[c]{@{}c@{}}Samples in  \\ Test Set\end{tabular}} \\ \hline
CEDAR\cite{11} & 1321 & 1329 & 1775 & 875 \\ \hline
Hindi-Dataset\cite{8} & 3840 & 4800 & 5788 & 2852 \\ \hline
Bengali-Dataset\cite{8} & 2400 & 3000 & 3618 & 1782 \\ \hline
SIGCOMP Dataset (Chinese)\cite{12} & 235 & 340 & 385 & 190 \\ \hline
SIGCOMP Dataset (Dutch)\cite{12} & 239 & 123 & 242 & 120 \\ \hline
\end{tabular}
\end{table}
\fi

\begin{table}[]
\centering
\caption{Benchmark Signature Datasets }
\begin{tabular}{|c|l|l|l|l|}
\hline
Dataset Name & \multicolumn{1}{c|}{\begin{tabular}[c]{@{}c@{}}Genuine\\ signatures\end{tabular}} & \multicolumn{1}{c|}{\begin{tabular}[c]{@{}c@{}}Fake\\ signatures\end{tabular}} & \multicolumn{1}{c|}{\begin{tabular}[c]{@{}c@{}}Samples in \\ Training Set\end{tabular}} & \multicolumn{1}{c|}{\begin{tabular}[c]{@{}c@{}}Samples in  \\ Test Set\end{tabular}} \\ \hline
CEDAR\cite{11} & 1321 & 1329 & 1775 & 875 \\ \hline
Hindi-Dataset\cite{8} & 3840 & 4800 & 5788 & 2852 \\ \hline
Bengali-Dataset\cite{8} & 2400 & 3000 & 3618 & 1782 \\ \hline
\end{tabular}

\end{table}

\subsection{Results \& Discussion}
We have evaluated our proposed model with various datasets mentioned in Table 1. We used the performance metric for accuracy of the model described in the paper \cite{9}. Performance with our proposed model and comparison with other state of the art method is described in Table 2. In case of Bengali \& Hindi Dataset, our proposed method outperformed other 3 state of the art methods but not be able to outperform Dutta et al., graph matching \& SigNet on CEDAR dataset. One possilbe reason for this result may be that for little amount of data, other unsupervised methods and Siamese network (trained on that dataset specifically) are better than ours.

\begin{table}[]
\centering
\caption{Comparison \& Performance of the proposed method with other state-of-the-art methods on mentioned signature datasets}
\begin{tabular}{|l|l|l|}
\hline
Dataset Name & Methods & Accuracy (\%) \\ \hline
CEDAR & \begin{tabular}[c]{@{}l@{}}Word Shape (GSC) (Kalera et al.\cite{2}) \\ Zernike moments (Chen and Srihari\cite{13}) \\ Graph matching (Chen and Srihari\cite{4}) \\ Surroundedness features (Kumar et al.\cite{12})\\ Dutta et al.\cite{5} \\ SigNet\cite{9}\\ Proposed Method\end{tabular} & \begin{tabular}[c]{@{}l@{}}78.50 \\ 83.60 \\ 92.10 \\ 91.67\\ \textbf{100}\\ \textbf{100}\\ 92\end{tabular} \\ \hline
Hindi Dataset & \begin{tabular}[c]{@{}l@{}}Pal et al.\cite{8} \\ Dutta et al.\cite{5}  \\ SigNet\cite{9}\\ Proposed Method\end{tabular} & \begin{tabular}[c]{@{}l@{}}75.53\\ 85.90\\ 84.64\\ \textbf{86}\end{tabular} \\ \hline
Bengali Dataset & \begin{tabular}[c]{@{}l@{}}Pal et al.\cite{8} \\ Dutta et al.\cite{5} \\ SigNet\cite{9}\\ Proposed Method\end{tabular} & \begin{tabular}[c]{@{}l@{}}66.18 \\ 84.90\\ 86.11\\ \textbf{94}\end{tabular} \\ \hline
\end{tabular}
\end{table}

\subsection{Conclusion}
In this paper, we have presented a framework based on ensemble learning for offline signature verification. Also, our experiments made on cross domain datasets indicate how well our ensemble model can robustly detect fraudulence of different handwriting style of different signers and forgers with diverse background and scripts. Furthermore, the Ensemble Model designed by us has surpassed the state-of-the-art results on most of the benchmark signature datasets from various scripts \& domain. Our future work in this problem will be focus on the multimodal model to incorporate both online \& offline signature in one model with different types of data augmentation.

\end{document}